\documentclass[conference]{IEEEtran}
\IEEEoverridecommandlockouts
\usepackage{cite}
\usepackage{amsmath,amssymb,amsfonts}
\usepackage{algorithmic}
\usepackage{graphicx}
\usepackage{textcomp}
\usepackage{xcolor}
\usepackage{multirow}
\usepackage{caption}
\usepackage{subcaption}
\usepackage{enumitem}
\usepackage{booktabs}
\usepackage{algorithmic}
\usepackage{algorithm}
\usepackage{hyperref}

\usepackage{xcolor}
\newcommand{\independent}{\perp\mkern-9.5mu\perp}

\usepackage{amsthm}

\newtheorem{assumption}{Assumption}[section]
\newtheorem{definition}{Definition}[section]
\newtheorem{theorem}{Theorem}[section]

\def\BibTeX{{\rm B\kern-.05em{\sc i\kern-.025em b}\kern-.08em
    T\kern-.1667em\lower.7ex\hbox{E}\kern-.125emX}}
\begin{document}

\title{Stochastic Intervention for Causal Effect Estimation\\
}


\author{\IEEEauthorblockN{Tri Dung Duong, Qian Li, Guandong Xu}
\IEEEauthorblockA{School of Computer Science,
University of Technology Sydney, Australia\\
\ TriDung.Duong@student.uts.edu.au} \{Qian.Li,\, Guandong.Xu\}@uts.edu.au}

\maketitle

\begin{abstract}
Causal inference methods are widely applied in various decision-making domains such as precision medicine, optimal policy and economics. Central to these applications is the treatment effect estimation of intervention strategies. Current estimation methods are mostly restricted to the deterministic treatment, which however, is unable to address the stochastic space treatment policies. Moreover, previous methods can only make binary yes-or-no decisions based on the treatment effect, lacking the capability of providing fine-grained effect estimation degree to explain the process of decision making.
In our study, we therefore advance the causal inference research to estimate stochastic intervention effect by devising a new stochastic propensity score and stochastic intervention effect estimator (SIE).
Meanwhile, we design a customized genetic algorithm specific to stochastic intervention effect (Ge-SIO) with the aim of providing causal evidence for decision making.
We provide the theoretical analysis and conduct an empirical study to justify that our proposed measures and algorithms can achieve a significant performance lift in comparison with state-of-the-art baselines. 
\end{abstract}


\section{Introduction}
    Causal inference increasingly plays a vitally important role in a wide range of fields including online marketing, precision medicine, political science, etc.
    For example, a typical concern in precision medicine is whether an \emph{alternative} medication treatment for a certain illness will lead to a better outcome \footnote{\emph{Treatment} and \emph{outcome} are terms in the theory of causal inference, which for example denote a promotion strategy taken and its resulting profit, respectively}. Treatment effect estimation can answer this question by comparing outcomes under different treatments.

    Estimating treatment effect is challenging, because only the factual outcome for a specific treatment assignment (say, treatment \texttt{A}) is observable, while the counterfactual outcome corresponding to alternative treatment \texttt{B} is usually unknown.
    Aiming at deriving the absent counterfactual outcomes, existing causal inference from observations methods can be categorized into these main branches: re-weighting methods \cite{gruber2011tmle, austin2015moving}, tree-based methods \cite{chipman2007bayesian, hill2011bayesian, wager2018estimation}, matching methods \cite{rosenbaum1983central, dehejia2002propensity,stuart2011matchit} and doubly robust learners \cite{econml, 10.5555/3104482.3104620}. 
    In general, the matching approaches focus on finding the comparable pairs based on distance metrics such as propensity score or Euclidean distance, while re-weighting methods assign each unit in the population a weight to equate groups based on the covariates. Meanwhile, tree-based machine learning models including decision tree or random forest are utilized in the tree-based approach to derive the counterfactual outcomes. Doubly Robust Learner is another recently developed approach that combines the propensity score weighting with the regression outcome to produce an unbiased and robust estimator. 
    

    
    Existing treatment effect estimation from observational data faces two major challenges. First, most of previous studies focus on the deterministic intervention which sets each individual a fixed treatment value, incapable of dealing with dynamic and stochastic intervention~\cite{dudik2014doubly, pearl2000models, tian2012identifying}.
    They can not address the question like ``how the health status changes (the desired outcome) for the patient if the doctor adopts 50\% dose reduction in the patient'', which might be of practical interest in real world.
    Second, existing methods fail in exploiting the relationships between the desired response and the intervention on the treatment, resulting in black-box effect estimation. 

    To address these issues, we propose a novel influence function based model to provide sufficient causal evidence to answer decision-making questions about stochastic interventions. Particularly, our model consists of three novel components: \emph{stochastic propensity score}, \emph{stochastic intervention effect estimator} (SIE) and \emph{customized genetic algorithm} for stochastic intervention optimization (Ge-SIO).
    The main contributions of our model are summarized below:

\begin{itemize}
    \item 
     We propose a new stochastic propensity score learning the treatment effect trajectory, which tackles the limitation of existing approaches only dealing with deterministic intervention effects.
    \item 
    Based on the general efficiency theory, we provide theoretical proof that SIE can achieve fast parametric convergence rates when the potential outcome model can not be perfectly estimated.   
    
    \item Ge-SIO is proposed to find the optimal intervention leading to the desired response, which can be widely applicable in domain-specific decision-making.
    

    
\end{itemize}



\section{Related works}
\label{section:related}
Conventionally, causal inference can be trickled by either the randomized experiment (also known as A/B testing in online settings) or observational data. 
    In randomized experiment, units are randomly assigned to a treatment and their responses are recorded. One treatment is selected as the best among the alternatives by comparing the predefined statistical criteria. While randomized experiments have been popular in traditional causal inference, it is prohibitively expensive \cite{chan2010evaluating, kohavi2011unexpected} and infeasible \cite{bottou2013counterfactual} in some real-world settings~\cite{li2017learning,wang2019polynomial, xu2020causality}.
    As an alternative method, observational study is becoming increasingly critical and available in many domains such as medicine, public policy and advertising. 
    However, observational study needs to deal with data absence problem, which differs fundamentally from supervised learning~\cite{}. This is simply because only the factual outcome (symptom) for a specific treatment assignment (say, treatment \texttt{A}) is observable, while the counterfactual outcome corresponding to alternative treatment \texttt{B} in the same situation is always unknown.
    
\subsection{Treatment Effect Estimation} 
The simplest way to estimate treatment effect in observational data is the matching method that finds the comparable units in the treated and controlled groups. The prominent matching methods include Propensity Score Matching (PSM) \cite{rosenbaum1983central, dehejia2002propensity} and Nearest Neightbor Matching (NNM) \cite{stuart2011matchit}. Particularly, for each treated individual, PSM and NNM select the nearest units in the controlled group based on some distance functions, and then calculate the difference between two paired outcomes. Another popular approach is reweighting method that involves in  building a classifier model to estimate the probability of a treatment assigned to a particular unit, and uses the predicted score as the weight for each unit in dataset. TMLE~\cite{gruber2011tmle} and IPSW~\cite{austin2015moving} fall into this category. Ordinary Linear Regression (OLS)~\cite{goldberger1964econometric} is another commonplace method that fits two linear regression models for the treated and controlled group, with each treatment as the input features, and the outcome as the output. The predicted counterfactual outcomes thereafter are used to calculate the treatment effect. Meanwhile, decision tree is a popular non-parametric machine learning model, attempting to build the decision rules for the regression and classification tasks. Bayesian Additive Regression Trees (BART) \cite{chipman2007bayesian, hill2011bayesian} and Causal Forest \cite{wager2018estimation} are the prominent tree-based method in causal inference. While BART~\cite{chipman2007bayesian, hill2011bayesian} builds the decision tree for the treated and controlled units, Causal Forest~\cite{wager2018estimation} constructs the Random Forest model to derive the counterfactual outcomes, and then calculates the difference between the paired potential outcomes to obtain the average treatment effect. They are proven to obtain the more accurate treatment effect than matching methods and reweighting methods in the non-linear outcome setting.

Doubly Robust Learner \cite{econml, 10.5555/3104482.3104620} is the recently proposed approach that constructs a regression estimator predicting the outcome based on the covariates and treatment, and builds a classifier model to fit the treatment. DRL finally combines the both predicted propensity score and predicted outcome to estimate treatment effect.

\subsection{Stochastic Intervention Optimization}
Our work connects to the uplift modelling which optimizes the treatment effect by uplifting the expected response under the treatment policy \cite{zaniewicz2013support, Personalized_Medicine, hansotia2002incremental, manahan2005proportional}. Uplift modelling measures the effectiveness of a treatment and then predicts the corresponding expected response. The most popular and widely-used approach is Separate Model Approach (SMA) \cite{zaniewicz2013support,Personalized_Medicine} which builds two different regression models. The first one uses treated unit data, whilst another works the controlled unit data. Several state-of-the-art machine learning models such as Random Forest, Gradient Boosting Regression or Adaboost can be used to construct the predictive model \cite{liaw2002classification, solomatine2004adaboost, friedman2001greedy}. The predicted responses are then calculated, and the optimal treatments are selected as the result. SMA has been widely applied in marketing \cite{hansotia2002incremental} and customer segmentation \cite{manahan2005proportional}. However, when dealing with the data containing a great deal of noisy and missing information, the model outcomes are prone to be incorrect and biased, which leads to the poor performance. Other commonplace methods include Class Transformation Model \cite{jaskowski2012uplift} and Uplift Random Forest \cite{guelman2014package} that build the classification model for each outcome in the dataset. These techniques therefore can only handle the categorical outcomes, instead the continuous ones.



\section{Preliminaries and Problem Definition}
\label{section:pre}
\subsection{Notation}

In this study, we consider the observational dataset $Z=\{\boldsymbol{x}_i,{y}_i,t_i\}_{i=1}^n$ with $n$ units, where $\boldsymbol{x}\in\mathbb{R}^{n\times d}$ is the $d$-dimensional covariate, 
$y$ and $t\in\{0,1\}$ are the outcome and the treatment for the unit, respectively.
The treatment variable is binary in many cases, thus the unit will be assigned to the control treatment if $t=0$, or the treated treatment if $t=1$. Accordingly, $y_0(\boldsymbol{x})$ and $y_1(\boldsymbol{x})$ are profit accrued from customer $i$ corresponding to either the controlled or treated group. The central goal of causal inference is to compare the potential outcomes of the same units under two or more treatment conditions, which is implemented by computing the average treatment effect (ATE), i.e.,
\begin{equation}
    \tau_{\text{ATE}}=\mathbb{E}[y_0(\boldsymbol{x})-y_1(\boldsymbol{x})]
    \label{eq:tau}
\end{equation}

\subsection{Propensity Score}
Rosenbaum and Rubin~\cite{rosenbaum1983central} first proposed propensity score technique to deal with the high-dimensional covariates. Particularly, propensity score can summarise the mechanism of treatment assignment and thus squeezes covariate space into one dimension to avoid the possible data sparseness issue~\cite{bang2005doubly, dehejia2002propensity, austin2015moving, hirano2003efficient}.  The propensity score is defined as the probability that a unit is assigned to a particular treatment $t=1$ given the covariate $\boldsymbol{x}$, i.e.,
\begin{equation}
    p_t(\boldsymbol{x}) = \mathbb{P}(t = 1 | \boldsymbol{x})
\end{equation}


In practice, one widely-adopted parametric model for propensity score $p_t(\boldsymbol{x})$ 
is the logistic regression
\begin{equation}
    \hat{p}_t(\boldsymbol{x})=\frac{1}{1+
\exp{(\boldsymbol{w}^{\top}\boldsymbol{x}+\omega_0)}}
\label{eq:ps}
\end{equation} 
where $\boldsymbol{w}$ and $\omega_0$ are estimated 
by minimizing the negative log-likelihood~\cite{martens2008systematic}. 
The propensity score is widely used in causal inference methods to estimate treatment effects from observational data~\cite{hirano2003efficient, pirracchio2016propensity, luo2010applying,abdia2017propensity}.

\subsection{Assumption}\label{sc:asp} Following the general practice in causal inference literature, the following two assumptions should be taken into consideration to ensure the identifiability of the treatment effect, i.e. \emph{Positivity} and \emph{Ignorability}. 

\begin{assumption} [Positivity]. Each unit has a positive probability to be assigned by a treatment, i.e.,
\begin{equation}
    p_t(\boldsymbol{x}) >0, \quad\forall \boldsymbol{x}\text{ and } t
\end{equation}
\end{assumption}



\begin{assumption}[Ignorability]
The assignment to the treatment $t$ is independent of the outcomes $\boldsymbol{y}$ given covariates $\boldsymbol{x}$
\begin{equation}
   y_1 , y_0  \independent t|\boldsymbol{x}
\end{equation}
\end{assumption}

\section{Stochastic Intervention Effect}
\label{section:estimation}




The stochastic intervention effect can be expressed by the difference between the observed outcome and the counterfactual outcome under the stochastic intervention. Because the observed outcome is fixed, stochastic intervention effect estimation is transformed as the problem of estimating the counterfactual outcome.

\subsection{Stochastic Counterfactual Outcome}
To estimate the counterfactual outcome, we first propose a flexible and task-specific stochastic propensity score to characterize the stochastic intervention.
 




\begin{definition}[Stochastic Propensity Score]{ The stochastic propensity score with respect to stochastic degree $\delta$ is 
\begin{equation}
    q_{t}( \boldsymbol{x},\delta) = \frac{\delta \cdot\hat{p}_t(\boldsymbol{x})}{\delta \cdot\hat{p}_t(\boldsymbol{x}) + 1 - \hat{p}_t(\boldsymbol{x})}
    \label{eq:incrps}
\end{equation}
where $\hat{p}_t(\boldsymbol{x})$ is denoted by
}
\begin{equation}
    \hat{p}_{t}(\boldsymbol{x})=\frac{\exp \left(\sum_{j=1}^{s} \beta_{j} g_{j}\left(\boldsymbol{x}\right)\right)}{1+\exp \left(\sum_{j=1}^{s} \beta_{j} g_{j}\left(\boldsymbol{x}\right)\right)}
    \label{eq:ourps}
\end{equation}
\label{def:ips}
where $\{ g_1, \cdots, g_s\}$ are nonlinear basis functions.
\end{definition}


The proposed stochastic propensity score in Definition \ref{def:ips} has two promising properties compared with \eqref{eq:ps}. 
On the one hand, propensity score \eqref{eq:ps} fails to quantify the causal effect under stochastic intervention. 
So we introduce $\delta$ in \eqref{eq:incrps} to represent the stochastic intervention indicating the extent to which the propensity scores are fluctuated from their actual observational values. 
For instance, the stochastic intervention that the doctor adopts 50\% dose increase in the patient can be expressed by $\delta = 1.5$. 

On the other hand, the linear term  $\boldsymbol{w}^{\top}\boldsymbol{x}+\omega_0$ in Eq.~\eqref{eq:ps} may lead to misspecification~\cite{dalessandro2012causally} if there are higher-order terms or non-linear trends among covariates $\boldsymbol{x}$.
So we propose to use a sum of nonlinear function $\sum_{j=1}^{s} \beta_{j} g_{j}$ in  \eqref{eq:ourps} that captures the non-linearity involving in covariates to create an unbiased estimator of treatment effect.

On the basis of the stochastic propensity score, we propose an influence function specific to estimate counterfactual outcome under stochastic intervention. 
Meanwhile, we also analyze the asymptotic behavior of the counterfactual outcome with theoretical guarantees. We prove that our influence function can achieve double robustness and fast parametric convergence rates. 


\begin{theorem}
\label{th:sto}
With the stochastic intervention of degree $\delta$ on observed data $z=(\boldsymbol{x},y,t)$, we have
\begin{equation}
    \varphi(z,\delta)=q_{t}(\boldsymbol{x},\delta)\cdot m_1(\boldsymbol{x},y)+(1-q_{t}(\boldsymbol{x},\delta))\cdot m_0(\boldsymbol{x},y)
    \label{eq:varphi}
\end{equation}
being the efficient influence function for the resulting counterfactual outcome $\hat{\psi}$, i.e., 
\begin{equation}
    \hat{\psi}
    =\mathbb{P}_{n}\left[\varphi(z,\delta)\right]
\label{eq:e_psi}
\end{equation}
where $m_1(\boldsymbol{x},y)$ or $m_0(\boldsymbol{x},y)$ is given by
\begin{equation}
    m_t(\boldsymbol{x},y)=\frac{\mathbb{I}_{t}\cdot({y-\hat{\mu}(\boldsymbol{x},t)})}{t\cdot\hat{p}_t(\boldsymbol{x})+(1-t)(1-\hat{p}_t(\boldsymbol{x}))}+\hat{\mu}(\boldsymbol{x},t)
    \label{eq:m}
\end{equation}
and $\mathbb{I}_{t}$ is an indicator function, $\hat{p}_t$ is the estimated propensity score in Eq.~\eqref{eq:ourps} and $\hat{\mu}$ is potential outcomes model that can be fitted by machine learning methods.
\end{theorem}



\begin{proof}


Throughout we assume the observed data quantity $\psi$ can be estimated under the positivity assumption from Section~\ref{sc:asp}.
For the unknown ground-truth $\psi(\delta)$, we will prove $\varphi$ is the influence function of $\psi(\delta)$ in Eq.~\eqref{eq:e_psi} by checking 
\begin{equation}
    \int \hat{\psi}(y,x,t,\mathbb{P}) d \mathbb{P}=\int\left(\varphi(y,x,t,\delta)-\psi \right) d \mathbb{P}=0
    \label{eq:property}
\end{equation} Eq.~\eqref{eq:property} indicates that the uncentered influence function $\varphi$ is unbiased for $\psi$. Given $q_t(\boldsymbol{x},\delta)$ as the stochastic propensity score in Eq.~\eqref{eq:incrps},
we check the property~\eqref{eq:property} by
\begin{equation*}
\small
\begin{split}
    &\int\left(\varphi(y,x,t,\delta)-\psi\right) d \mathbb{P}\\
    &=\int\left\{q_t\cdot m_1(\boldsymbol{x},y)+(1-q_t)m_0(\boldsymbol{x},y)-\psi(\delta)\right\}d\mathbb{P}(y,x,t,\delta)\\
    &=\int\{q_t\frac{\mathbb{I}_{t=1}\cdot(y-\hat{\mu}(x,1))}{\hat{p}_t}+(1-q_t)\frac{\mathbb{I}_{t=0}\cdot(y-\hat{\mu}(x,0))}{1-\hat{p}_t}\\ &+q_t\hat{\mu}(x,1)+(1-q_t)\hat{\mu}(x,0)-\psi(\delta)\}d\mathbb{P}(y,x,t,\delta)\\
    &=\int\{q_t\frac{\mathbb{I}_{t=1}\cdot(y-\hat{\mu}(x,1))}{\hat{p}_t}+(1-q_t)\frac{\mathbb{I}_{t=0}\cdot(y-\hat{\mu}(x,0))}{1-\hat{p}_t} \\ &+q_t\hat{\mu}(x,1)+(1-q_t)\hat{\mu}(x,0)-\mathbb{E}[q_t\hat{\mu}(x,1)\\
    &+(1-q_t)\hat{\mu}(x,0)]\}d\mathbb{P}(y,x,t,\delta)\\
    &\overset{(1)}{=}\int\left\{q_t\frac{\mathbb{I}_{t=1}\cdot(y-\hat{\mu}(x,1))}{\hat{p}_t}\right\}d\mathbb{P}(y,x,t,\delta)\\ &+\int\left\{(1-q_t)\frac{\mathbb{I}_{t=0}\cdot(y-\hat{\mu}(x,0))}{1-\hat{p}_t}\right\}d\mathbb{P}(y,x,t,\delta)\\
    &=\int\left\{q_t\frac{\mathbb{I}_{t=1}\cdot y}{\hat{p}_t}+(1-q_t)\frac{\mathbb{I}_{t=0}\cdot y}{1-\hat{p}_t}\right\}d\mathbb{P}(y,x,t,\delta)\\
    &-\int\left\{q_t\frac{\mathbb{I}_{t=1}\cdot\hat{\mu}(x,1)}{\hat{p}_t}-(1-q_t)\frac{\mathbb{I}_{t=0}\cdot\hat{\hat{\mu}}(x,0)}{1-\hat{p}_t}\right\}d\mathbb{P}(x,t,\delta)\\
    &\overset{(2)}{=}0
\end{split}
\end{equation*}
The second equation (1) follows from the iterated expectation, and the second equation (2) follows from the definition of $\hat{\mu}(\boldsymbol{x},t)$ and the usual properties of conditional distribution $d\mathbb{P}(x,y,\delta)=d\mathbb{P}(y|x,\delta)d\mathbb{P}(x,\delta)$.
So far we have proved that $\varphi$ is the influence function of average treatment effect $\psi(\delta)$. 
We have proved that the uncentered efficient influence function can be used to construct unbiased semiparametric estimator for $\psi(\delta)$, i.e., that 
$\int \varphi\mathbb{P}=\psi$.
\end{proof}
\begin{algorithm}[!htbp]
\small
\caption{SIE: Stochastic Intervention Effect}
\label{alg:avg}
\begin{algorithmic}[1]
\renewcommand{\algorithmicrequire}{\textbf{Input:}}
\renewcommand{\algorithmicensure}{\textbf{Output:}}
\REQUIRE Observed units $\{z_i:(\boldsymbol{x}_i,t_i,y_i)\}_{i=1}^{n}$
\STATE Initialize a stochastic degree $\delta$. 
\STATE Randomly split $Z$ into $k$ disjoint groups
\WHILE{each group}
\STATE Fit the propensity score $\hat{p}_t(\boldsymbol{x}_i)$ by Eq.~\eqref{eq:ourps} 

\STATE Fit the potential outcome model $\hat{\mu}(\boldsymbol{x}_i,t_i)$
\STATE Compute $\tau_{i}=\hat{p}_{t}(\boldsymbol{x}_i) \hat{\mu}(\boldsymbol{x}_i,1)+\left(1-\hat{p}_{t}(\boldsymbol{x}_i)\right) \hat{\mu}(\boldsymbol{x}_i,0)$
\STATE Calculate $q_{t}( \boldsymbol{x}_i;\delta)$ by Eq.~\eqref{eq:incrps}
\STATE Calculate $m_{1}(\boldsymbol{x}_i)$ and $m_{0}(\boldsymbol{x}_i)$ by Eq.~\eqref{eq:m}
\STATE Calculate the influence function $\varphi(z_i,\delta)$ by Eq.~\eqref{eq:e_psi}.
\ENDWHILE
\STATE Compute $\hat{\tau}_{\text{ATE}}=\frac{1}{n} \sum_{i=1}^{n}\tau_{i}$
\STATE Compute $\hat{\tau}_{\text{SIE}}=\frac{1}{n} \sum_{i=1}^{n}(\varphi(z_i,\delta)-y_i)$

\ENSURE stochastic intervention effect $\tau_{\text{SIE}}$, ATE $\tau_{\text{ATE}}$
\end{algorithmic}
\end{algorithm}

\section{Stochastic Intervention Optimization}
Estimating the stochastic intervention effect is not enough, we are more interested in ``what is the optimal level/degree of treatment for a patient to achieve the most expected outcome?''. 
In this section, we apply influence-based estimator to search for the optimal intervention  
that achieves the optimal expected response over the whole population. 
We model the stochastic intervention using the stochastic propensity score $\hat{q}_t(\boldsymbol{x},\delta)$, and 
look for a set of stochastic interventions $\Delta=\{\delta_1^*,\cdots,\delta_n^*\}$ where the $i$-th intervention $\delta_i^*\in\Delta$ maximizes the expected response specific to $i$-th unit $z_i=(\boldsymbol{x}_i,y_i,t_i)$, denoted by $\varphi(z_i,\delta_i)$:
\begin{equation}
    \delta_i^*=\arg\max_{\delta_i}\varphi(z_i,\delta_i)
    \label{eq:delta}
 \end{equation}
\label{section:strategy}

 

Note that the optimization problem in Eq.~\eqref{eq:delta} is non-differentiable.
To avoid using further assumptions for solving it, we formulate a customised genetic algorithm~\cite{whitley1994genetic} (Ge-SIO) to exploit the search space in an efficient and flexible manner. 
The main advantage of Ge-SIO is model-agnostic which can handle with any black-box functions and data type. Therefore, with modifications specific to the intervention effect estimation, Ge-SIO solves Eq.~\eqref{eq:delta} through a process of natural selection. The input of Ge-SIO is the fitness function $\hat{\psi}(\cdot)$ and intervention regime $\Delta$.

For stochastic intervention optimization, each candidate solution is described by the $n$-dimensional intervention  $\Delta$  (the ``genes'') and the objective values of the candidates are evaluated by Eq~\eqref{eq:delta}. Usually, a random population of solutions is initialized, which undergoes through the process of evolution to obtain the better fitness function until the stopping condition is reached.
Specifically, Ge-SIO first selects $m$ solutions as the population of parents based on their fitness values. 
Among the selected parent solutions, $m$ solutions are chosen pairwise with the uniform distribution to produce children, which is called crossover process. The $n$-dimensional $\Delta$ are recombined by the simulated binary crossover recombinator. Crossover takes $m$ selected parents and combines them, for the sake of diversity to the solutions. 
The children, which constitute solutions, are modified by the mutation operator. Mutation has a small chance to change $\Delta$, which may create more fitter solutions. 
Thus, the Ge-SIO first generates children by crossover and modifies them by mutation thereafter. After the process of evolution is done, the fittest $\Delta$ is returned as the optimal solution to the desired expected response $\hat{\psi}$.
We run it with the number of generations to repeat the above process so as to find the optimal solution. The full stochastic intervention optimization algorithm is shown in Algorithm~\ref{alg:IEO}. 

\begin{algorithm}[H]
\small
\caption{Ge-SIO: Stochastic Intervention Optimization}
\label{alg:IEO}
\begin{algorithmic}[1]
\renewcommand{\algorithmicrequire}{\textbf{Input:}}
\renewcommand{\algorithmicensure}{\textbf{Output:}}
\REQUIRE Observed units $\{z_i:(\boldsymbol{x}_i,t_i,y_i)\}_{i=1}^{n}$
\STATE Initialize a batch of population $\Gamma=\{\boldsymbol{\Delta}_1,\cdots,\boldsymbol{\Delta}_m\}$ with $\boldsymbol{\Delta}_i\sim\mathcal{N}(\boldsymbol{\mu},\boldsymbol{\nu})$

\FOR{$G$ generation}
\FOR{$k=1,\cdots,m$}
\FOR{$i=1,\cdots,n$}
\STATE Compute $q_t(\boldsymbol{x},\delta_i)$ by Eq.~\eqref{eq:incrps}
\STATE Calculate $m_{1}(\boldsymbol{x}_i)$ and $m_{0}(\boldsymbol{x}_i)$ by Eq.~\eqref{eq:m}
\STATE Calculate $\varphi(z_i,\delta)$ by Eq.~\eqref{eq:varphi}. 
\ENDFOR
\STATE Compute $k$-th fitness  $\Phi(\Delta_k)=\sum_{i=1}^{n}\varphi(z_i,\delta)$ 
\ENDFOR
\STATE Select $\Delta_1,\cdots,\Delta_m\in\Gamma$ based on its fitness function
\STATE Randomly pair $\lceil m/2\rceil$  $\{\Delta_1,\Delta_2\}\in\Gamma$
\FOR{each pair $\{\Delta_1,\Delta_2\}$}
\STATE Perform uniform $\text{crossover}(\Delta_1,\Delta_2)\rightarrow \Delta_1^{\prime},\Delta_2^{\prime}$
\STATE Perform uniform mutation  $\Delta_1^{\prime}\rightarrow \tilde{\Delta}_1,\Delta_2^{\prime}\rightarrow \tilde{\Delta}_2$ 
\STATE Update $\Gamma$ by replacing $\{\Delta_1,\Delta_2\}$ with $\{\tilde{\Delta}_1,\tilde{\Delta}_2\}$
\ENDFOR


\ENDFOR
\STATE Choose $\Delta^*=\arg\max_{\Delta\in\Gamma}\,\Phi(\boldsymbol{\Delta})$ 
\ENSURE $\Delta^*$
\end{algorithmic}
\end{algorithm}

\section{Experiments and Results}
\label{section:experiment}
In this section, we conduct intensive experiments and compare our methods with state-of-the-art methods on two tasks: average treatment effect estimation and stochastic intervention effect optimization. Recall that the influence-based estimator $\varphi$ depends on the nuisance function of propensity score $p_t$ and outcome $\mu$. We first perform average treatment effect estimation to confirm that $\hat{p}_t$ and $\hat{\mu}$ are unbiased and robust estimators. 
Moreover, the stochastic intervention optimization task is carried out to demonstrate the effectiveness of our Ge-SIO, as well as investigate the impact of stochastic parameter $\delta$ on the expected response.


\subsection{Baselines}
\label{sec:base}

We briefly describe the comparison methods which are used in two tasks of treatment effect estimation and stochastic intervention optimization.

\subsubsection{Treatment effect estimation}
We can not able to directly evaluate SIE on the estimation of stochastic intervention effect, because no dataset with ground-truth stochastic counterfactual outcome is available. On the contrary, the benchmark datasets having two potential outcomes are available for ATE estimation. Therefore, we perform ATE estimation to evaluate the robustness of $\hat{p}_t$ and $\hat{\mu}$ thus to indirectly evaluate the performance of SIE.
We use Gradient Boosting Regression with 100 regressors for the potential outcome models $\hat{\mu}$.
We compare our proposed estimator (SIE) with the following baselines including Doubly Robust Leaner~\cite{10.5555/3104482.3104620}, IPWE~\cite{austin2015moving}, BART~\cite{hill2011bayesian}, Causal Forest~\cite{wager2018estimation, athey2019generalized}, TMLE~\cite{gruber2011tmle} and OLS~\cite{goldberger1964econometric}. 
Regarding implementation and parameters setup, we adopt Causal Forest~\cite{wager2018estimation, athey2019generalized} with 100 trees, BART \cite{hill2011bayesian} with 50 trees and TMLE~\cite{gruber2011tmle} from the libraries of cforest, pybart and zepid in Python. For Doubly Robust Learner (DR) \cite{10.5555/3104482.3104620}, we use the two implementations, i.e. LinearDR and ForestDR from the package EconML~\cite{econml} with Gradient Boosting Regressor with 100 regressors as the regression model, and Gradient Boosting Classifier with 200 regressors as the propensity score model. Ultimately, we use package DoWhy \cite{sharma2020dowhy} for IPWE \cite{austin2015moving} and OLS.

\subsubsection{Stochastic Intervention Optimization} We compare our proposed method (Ge-SIO) with Separate Model Approach (SMA) with different settings. SMA \cite{zaniewicz2013support,Personalized_Medicine} aims to build two separate regression models for the outcome prediction in the treated and controlled group, respectively. Under the setting of SMA, we apply four well-known models for predicting outcome including Random Forest (SMA-RF) \cite{soltys2015ensemble, grimmer2017estimating}, Gradient Boosting Regressor (SMA-GBR) \cite{friedman2001greedy}, Support Vector Regressor (SMA-SVR) \cite{zaniewicz2013support}, and AdaBoost (SMA-AB) \cite{solomatine2004adaboost}. We also compare the performance of these models with the random policy to justify that optimization algorithms can help to target the potential customers to generate greater revenue. 
For the settings of SMA, we use Gradient Boosting Regressor with 1000 regressors, AdaBoost Regression with 50 regressors, and Random Forest Tree Regressor with 100 trees.

\subsection{Datasets} 
\texttt{IHDP} \cite{hill2011bayesian} is a standard semi-synthetic dataset used in the \emph{Infant Health and Development Program}, which is a popularly used semi-synthetic benchmark containing both the factual and counterfactual outcomes. We conduct the experiment on 100 simulations of \texttt{IHDP} dataset, in which each dataset is divided into training and testing set. The training dataset is highly imbalanced with 139 treated and 608 controlled units out of total 747 units, respectively, whilst the testing dataset has 75 units. Each unit has 25 covariates representing the individuals' characteristics. The outcomes are their IQ scores at age 3~\cite{dorie2016npci}. 


Online promotion dataset (\texttt{OP} Dataset) provided by EconML project \cite{econml} is chosen to evalute stochastic intervention optimization \footnote{\url{https://msalicedatapublic.blob.core.windows.net/datasets/Pricing/pricing\_sample.csv}}. This dataset consists of 10k records in online marketing scenario with the treatment of discount price and the outcome of revenue, each represents a customer with 11 covariates. We split the data into two part: 80\% for training and 20\% for testing set. We run 100 repeated experiments with different random states to ensure the model outcome reliability. With this dataset, we aim to investigate how different price policies applied to different customers will result in the best generated revenue. We directly model the revenue as the expected response for the uplift modelling algorithm.

\subsection{Evaluation Metrics}
In this section, we briefly describe the two evaluation metrics used for treatment effect estimation and optimization.
Based on Eq.~\eqref{eq:tau}, we define the metric for evaluating the task of treatment effect estimation as the mean absolute error between the estimated and true ATE: 
\begin{equation}
    \epsilon_{ATE} = |\hat{\tau}_{\text{ATE}} - \tau_{\text{ATE}}|
\end{equation}
Moreover, the main performance metric in the  task is the expected value of the response under the proposed treatment, followed by the uplifting models study~\cite{zhao2017uplift, hitsch2018heterogeneous}. 




\subsection{Results and Discussions}

In this section, we aim to report the experimental results of 1) how our proposed estimator (SIE) can accurately estimate the average treatment effect; 2) how our optimization algorithm (Ge-SIO) can be used for finding optimal stochastic intervention in online promotion application; and 3) how the impacts of data size and stochastic degree are. 

\subsubsection{Treatment Effect Estimation}
The results of $\epsilon_{ATE}$ derived from \texttt{IHDP} dataset with 100 simulations and \texttt{OP} dataset with 100 repeated experiments are presented in the Table~\ref{table:ihdp} and Table~\ref{table:customer}, respectively. As seen clearly, amongst all approaches, our proposed method SIE achieves the best performance of the estimated ATE, while the Doubly Robust Learner performs next satisfactorily. Particularly, on \texttt{IHDP}, SIE outperforms all other methods in both training and testing set. In order to investigate the impact of data size chosen on estimation, we also run experiments and plot the performance of models in different data sizes in Figure~\ref{fig:ihdp}. Notably, SIE consistently produces the more accurate average treatment effect than others as the data size increases. Causal Forest and Doubly Robust Learner also produce the very competitive results, whereas the lowest performance belongs to IPWE. Turning to the experimental results on online promotion dataset in Table~\ref{table:customer}, SIE also has an outstanding performance consistently. Additionally, Doubly Robust Learner methods are ranked second, while the competitive results are recorded with BART. It is also worthy to note that although TMLE performs well in training set, its performance likely degrades when dealing with out-of-sample data in testing set. Overall, these results validate that our proposed SIE estimator proves to be effective and has an outstanding performance in the small and highly imbalanced dataset (\texttt{IHDP}) as well as in real-world application dataset (\texttt{OP}). 

\begin{table}[!htb]	
	\centering
	\small
	\caption{$\epsilon_{ATE}$ on 100 simulations of \texttt{IHDP} for training and testing  (lower is better).}
	\begin{tabular}{ccc}
\hline
		\multirow{2}{*}{Method} &\multicolumn{2}{c}{\texttt{IHDP Dataset} ($\epsilon_{\mathrm{ATE}}\pm\texttt{std}$)}
		\\ \cmidrule{2-3}
		&     Train         &         Test  \\
\hline
		OLS & 0.746 $\pm$ 0.140 &1.264 $\pm$ 0.250 \\
BART & 1.087 $\pm$ 0.120 & 2.808 $\pm$ 0.100 \\
Causal Forest & 0.360 $\pm$ 0.050 & 0.883  $\pm$ 0.614\\
TMLE & 0.326 $\pm$ 0.060 & 0.831 $\pm$ 1.750 \\
ForestDRLearner & 1.044 $\pm$ 0.040 & 1.224 $\pm$ 0.080\\
LinearDRLearner & 0.691 $\pm$ 0.080 & 0.797 $\pm$ 0.170  \\
IPWE & 1.701 $\pm$ 0.140 &  5.897  $\pm$ 0.300 \\
\hline
SIE & \textbf{0.284 $\pm$ 0.050} & \textbf{0.424 $\pm$ 0.090 } \\
\hline

	\end{tabular}
	\label{table:ihdp}
\end{table} 

\begin{figure}[!htb]
\label{fig:ihdp}
\centerline{\includegraphics[width=0.46\textwidth]{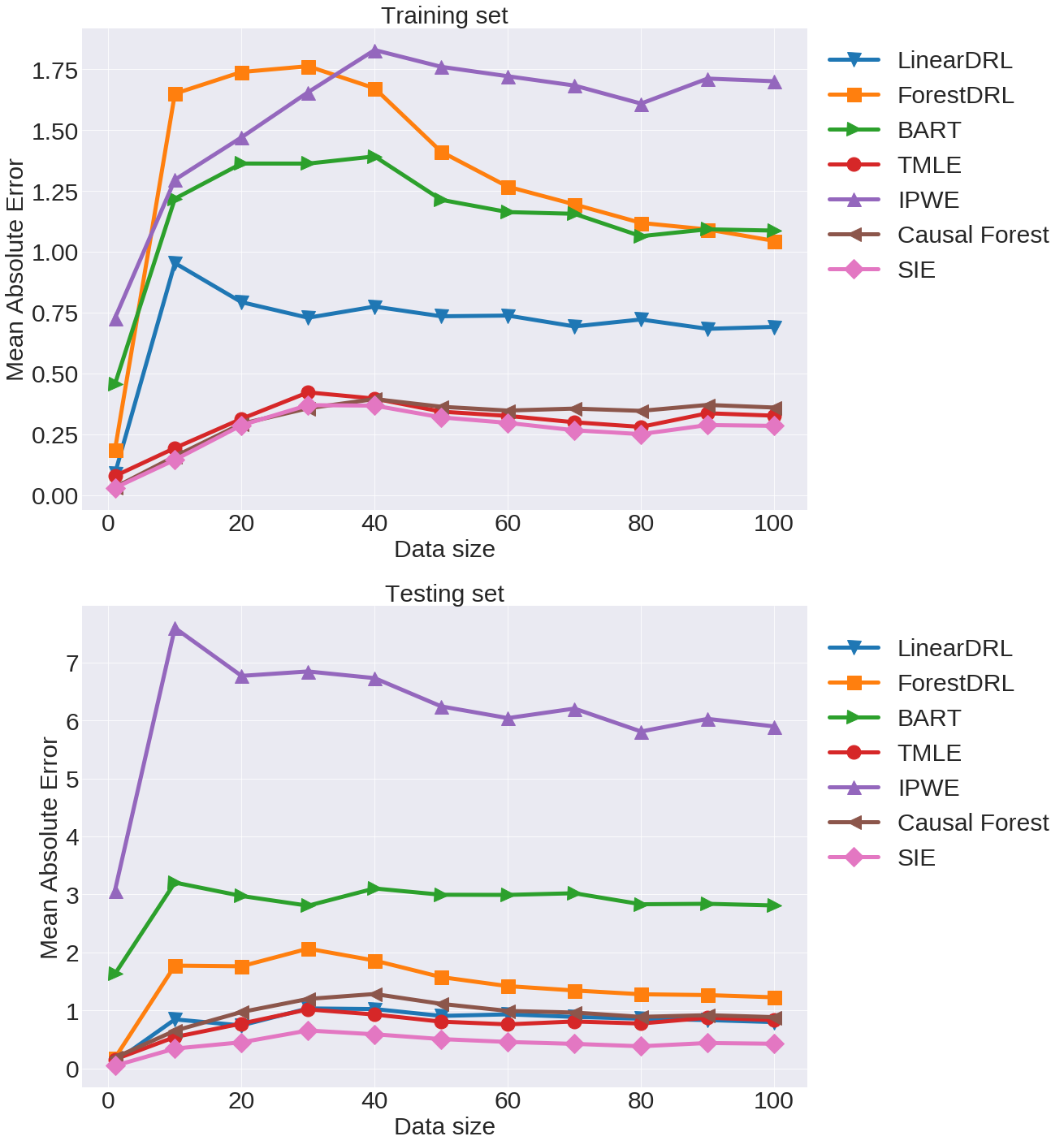}}
\caption{$\epsilon_{ATE}$ on \texttt{IHDP} under different datasize}
\label{fig:ihdp}
\end{figure}

\begin{table}[!htb]	
	\centering
	\small
	\caption{$\epsilon_{ATE}$ on \texttt{OP} dataset in 100 repeated experiments (lower is better). }
	\begin{tabular}{ccc}
\hline
		\multirow{2}{*}{Method} &\multicolumn{2}{c}{\texttt{OP Dataset}  ($\epsilon_{\mathrm{ATE}}\pm \texttt{std}$)}
		\\ \cmidrule{2-3}
		&     Train         &         Test  \\
\hline
                          OLS &   5.906 $\pm$ 0.004 &   5.907 $\pm$ 0.000 \\
                        BART &   0.504 $\pm$ 0.042 &   0.505 $\pm$ 0.043 \\
                Causal Forest &    3.520 $\pm$ 0.034 &    3.520 $\pm$ 0.034 \\
                         TMLE &   0.660 $\pm$ 0.000 &   3.273 $\pm$ 0.000 \\
              ForestDRLearner &    0.240 $\pm$ 0.014 &   0.241 $\pm$ 0.013 \\
              LinearDRLearner &    0.139 $\pm$ 0.009 &    0.139 $\pm$ 0.008 \\
                         IPWE &   5.908 $\pm$ 0.004 &   5.908 $\pm$ 0.015 \\
\hline
SIE &     \textbf{0.137 $\pm$ 0.000} &   \textbf{0.119 $\pm$ 0.000} \\
\hline

	\end{tabular}
	\label{table:customer}
\end{table}

\subsubsection{Stochastic Intervention Optimization}
\begin{figure}[!htb]
\centerline{\includegraphics[width=0.44\textwidth]{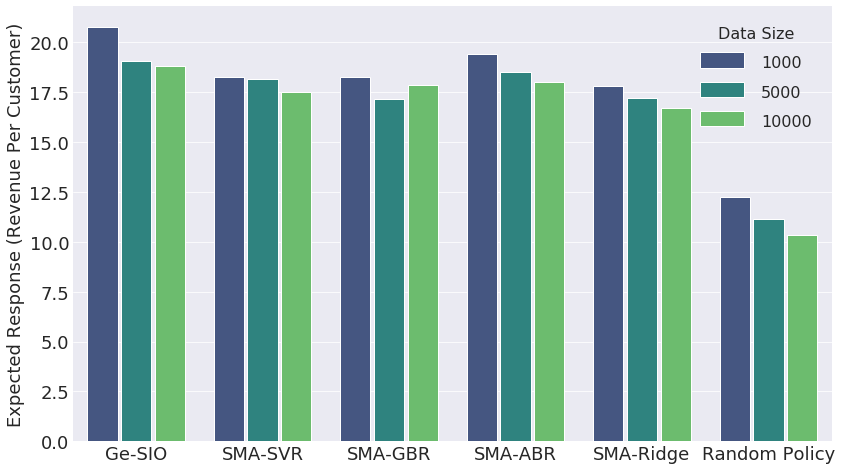}}
\caption{Expected revenue per customer from \texttt{OP} dataset by different models}
\label{fig:revenue}
\end{figure}

For the online promotion scenario, we model the revenue in dataset as the expected response of each customer under proposed treatment. Figure~\ref{fig:revenue} presents the revenue of uplifting modeling methods with different data sizes including 1000, 5000 and 10000 records. We set 100 generations for our Ge-SIO. Apparently, Ge-SIO generally produces the greatest revenue in all three datasizes, while SMA-ABR achieves the second-best performance with a very competitive result. Moreover, there is no significant difference in the performance of SMA with different settings. In contrast, the lowest revenue is generated by the random stochastic intervention that fails to choose the target customers to provide the promotion. The possible reason behind our proposed method's outstanding performance is that instead of getting the uplift signal like SMA, we directly intervene into the propensity score to produce the best stochastic intervention. From the business view, this emphasizes the crucial importance of the stochastic intervention optimization in online marketing campaign. 

\begin{figure}[!htb]
\centerline{\includegraphics[width=0.5\textwidth, scale=2]{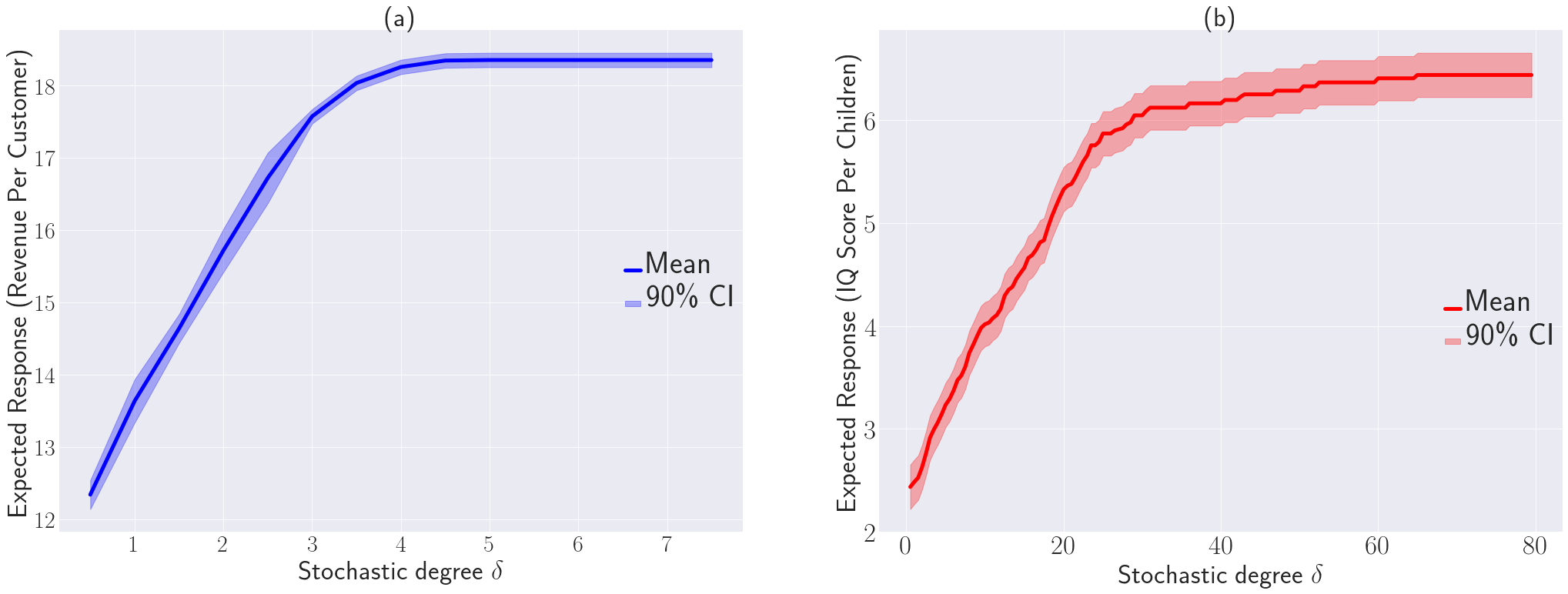}}
\caption{(a) Expected revenue per customer from \texttt{OP} dataset with uniform 90\% confidence. (b) Expected IQ score per children from \texttt{IHDP} dataset with uniform 90\% confidence}
\label{fig:variation}
\end{figure}

On the other hand, Figure~\ref{fig:variation} provides the information on the expected response with the various stochastic degree $\delta$ in \texttt{OP} and \texttt{IHDP} dataset with 90\% confidence interval. More specifically, when increasing degree $\delta$ from 0 to 5, the expected revenue also increases accordingly. The revenue thereafter reaches the highest point and remains nearly stable when $\delta$ is greater than 5. Similarly, the expected IQ score per children in the \texttt{IHDP} dataset also witnesses the same trend: the IQ score climbs gradually as stochastic degree $\delta$ rises. The plot of the relationship between the expected response and stochastic degree $\delta$ provides valuable insights into the degree of intervention we should make to achieve the optimal stochastic intervention, which can greatly facilitate the decision-making process.

\section{Conclusion}
\label{section:conclusion}
Causal inference increasingly gains the attention from both academia and industry as a powerful tool to deal with the scenario where people are not only interested to know the treatment effect but also the optimal intervention for the expected responses~\cite{wang2020joint,yin2021leveraging}. 
To extend causal inference to addressing stochastic interventions, this paper focuses on the dynamic intervention that is not discussed much in the recent study. In general, the contribution of this study is twofold. Based on stochastic propensity score, we propose a novel stochastic intervention effect estimator along with a customised genetic algorithm for stochastic intervention optimization. Our method can learn the trajectory of the stochastic intervention effect, providing causal insights for decision-making applications. 
Theoretical and numerical results justify that our methods outperform state-of-the-art baselines in both treatment effect estimation and stochastic intervention optimization. 

\section*{Acknowledgement}
This work is partially supported by the Australian Research Council (ARC) under Grant No. DP200101374 and LP170100891.

\bibliographystyle{unsrt}
\bibliography{sample-base}

\end{document}